\documentclass[10pt,twocolumn,letterpaper]{article}

\usepackage{iccv}
\usepackage{times}
\usepackage{epsfig}
\usepackage{graphicx}
\usepackage{amsmath}
\usepackage{amssymb}
\usepackage{multirow}
\usepackage{subcaption}

\iccvfinalcopy 

\def\iccvPaperID{2654} 

\ificcvfinal\fi
\begin{document}

\title{Deeper, Broader and Artier Domain Generalization}

\author{Da Li \quad Yongxin Yang \quad Yi-Zhe Song \quad Timothy M. Hospedales\\
Queen Mary University of London \quad
University of Edinburgh \\
{\tt\small \{da.li, yongxin.yang, yizhe.song\}@qmul.ac.uk, 
t.hospedales@ed.ac.uk}
}

\maketitle
\thispagestyle{empty}

\begin{abstract}
The problem of domain generalization is to learn from multiple training domains, and extract a domain-agnostic model that can then be applied to an unseen domain. Domain generalization (DG) has a clear motivation in contexts where there are target domains with distinct characteristics, yet sparse data for training. For example recognition in sketch images, which are distinctly more abstract and rarer than photos. Nevertheless, DG methods have primarily been evaluated on photo-only benchmarks focusing on alleviating the dataset bias where both problems of domain distinctiveness and data sparsity can be minimal. We argue that these benchmarks are overly straightforward, and show that simple deep learning baselines perform surprisingly well on them.

In this paper, we make two main contributions: Firstly, we build upon the favorable domain shift-robust properties of deep learning methods, and develop a low-rank parameterized CNN model for end-to-end DG learning. Secondly, we develop a DG benchmark dataset covering photo, sketch, cartoon and painting domains. This is both more practically relevant, and harder (bigger domain shift) than existing benchmarks. The results show that our method outperforms existing DG alternatives, and our dataset provides a more significant DG challenge to drive future research.
\end{abstract}

\section{Introduction}
Learning models that can bridge train-test domain-shift is a topical issue in computer vision and beyond. In vision this has been motivated recently by the observation of significant bias across popular datasets \cite{torralba2011dataset_bias}, and the poor performance of state-of-the-art models when applied across datasets. Existing approaches can broadly be categorized into domain \emph{adaptation} (DA) methods, that use (un)labeled target data to adapt source model(s) to a specific target domain \cite{saenko2010domainAdapt}; and domain \emph{generalization} (DG) approaches, that learn a domain agnostic model from multiple sources that can be applied to any target domain \cite{ECCV12_Khosla,ghifary2015domain}. While DA has been more commonly studied,  DG is the more valuable yet challenging setting, as it does not require acquisition of a large target domain set for off-line analysis to drive adaptation. Such data may not even exist if the target domain is sparse. Instead it aims to produce a more human-like model, where there is a deeper semantic sharing across different domains -- a dog is a dog no matter if it is depicted in the form of a photo, cartoon, painting, or indeed, a sketch. 

\begin{figure}[t]
%
%
\includegraphics[width=1.0\columnwidth]{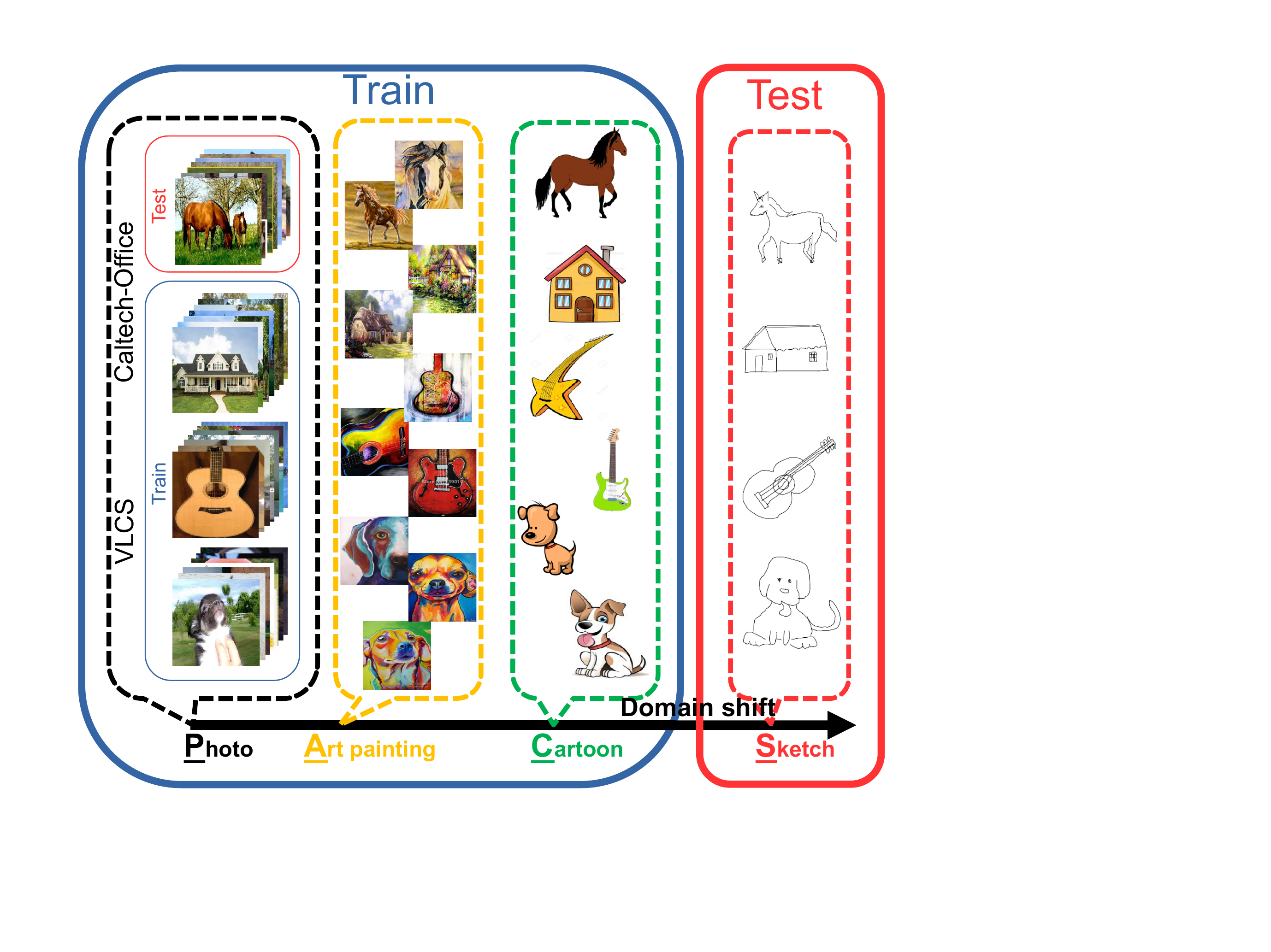}
\caption{Contrast between prior Caltech Office and VLCS datasets versus our new PACS dataset. The domain generalization task is to recognize categories in an unseen testing domain. PACS provides more diverse domains with bigger more challenging domain-shifts between them. }\label{fig:datasetSummary}
\end{figure}

The most popular existing DA/DG benchmarks define domains as photos of objects spanning different camera types \cite{saenko2010domainAdapt}, or datasets collected with different composition biases \cite{torralba2011dataset_bias}. While these benchmarks provide a good start, we argue that they are neither well motivated nor hard enough to drive the field. \emph{Motivation}: The constituent domains/datasets in existing benchmarks are based upon conventional photos, albeit with different camera types or composition bias. However there exist enough photos, that one could in principle collect enough target domain-specific data to train a good model, or enough diverse data to cover all domains and minimize bias 
(thus negating the need for DA). A more compelling motivation is domains where the total available images is fundamentally constrained, such as for particular styles of art \cite{crowley2016artDetection,wu2014matchAcrossStyle}, and sketches \cite{yu2016sketchShoe,eitz2012howToSketch,yu2017sketch,song2017sketch}. Compared to photos, there may simply not be enough examples of a given art style to train a good model, even if we are willing to spend the effort. \emph{Difficulty:} The camera type and bias differences between domains in existing benchmarks are already partially bridged by contemporary Deep features \cite{donahue2014decaf,yosinski2014howTransferable}, thus questioning the need for DA or DG methods. In this paper, we show that multi-domain deep learning provides a very simple but highly effective approach to DG that outperforms existing purpose-designed methods.

To address these limitations, we provide a harder and better motivated benchmark dataset PACS, consisting of images from photo (P), art painting (A), cartoon (C), and sketch (S) domains. This benchmark carries two important advancements over prior examples: (i) it extends the previously photo-only setting in DA/DG research, and uniquely includes domains that are maximally distinct from each other, spanning a wide spectrum of visual abstraction, from photos that are the least abstract to human sketches which are the most abstract; (ii) it is more reflective of a real-world task where a target domain (such as sketch) is intrinsically sparse, and so DG from a more abundant domain (such as photos) is really necessary. As illustrated qualitatively in Fig.~\ref{fig:datasetSummary}, the benchmark is harder, as the domains are visually more distinct than in prior datasets. We explore these differences quantitatively in Sec.~\ref{sub:datasetAnalysis}.


There have been a variety of prior approaches to DG based on SVM \cite{ECCV12_Khosla,xu2014exploiting}, subspace learning \cite{muandet2013domainGen}, metric learning \cite{fang2013unbiased}, and autoencoders \cite{ghifary2015domain}. Despite their differences, most of these have looked at fixed shallow features. In this paper, we address the question of how end-to-end learning of deep features impacts the DG setting. Our deep learning approach trains on multiple source domains, and extracts \emph{both} domain agnostic features (e.g., convolutional kernels), and classifier (e.g., final FC layer) for transfer to a new target domain. This approach can be seen as a deep multi-class generalization of the shallow binary {Undo Bias} method \cite{ECCV12_Khosla}, which takes the form of a dynamically parameterized deep neural network \cite{sigaud2015gatedInventory}. However, the resulting number of parameters grows linearly with the number of source domains (of which ultimately, we expect many for DG), increasing overfitting risk. To address this we develop a low-rank parameterized neural network which reduces the number of parameters. Furthermore the low-rank approach provides an additional route to knowledge sharing besides through explicit parameterization. In particular it has the further benefit of automatically modeling how related the different domains are (e.g., perhaps sketch is similar to cartoon; and cartoon is similar to painting), and also how the degree of sharing should vary at each layer of the CNN.
%
%

To summarize our contributions: Firstly, we highlight the weaknesses of existing methods (they lose to a simple deep learning baseline) and datasets (their domain shift is small). Second, we introduce a new, better motivated, and more challenging DG benchmark. Finally, we develop a novel DG method based on low-rank parameterized CNNs that shows favorable performance compared to prior work.



\section{Related work}

\vspace{0.1cm}\noindent\textbf{Domain Generalization}\quad
Despite different methodological tools (SVM, subspace learning, autoencoders, etc), existing methods approach DG based on a few different intuitions. One is to project the data to a new domain invariant representation where the differences between training domains is minimized \cite{muandet2013domainGen,ghifary2015domain}, with the intuition that such a space will also be good for an unseen testing domain. Another intuition is to predict which known domain a testing sample seems most relevant to, and use that classifier \cite{xu2014exploiting}. Finally, there is the idea of generating a domain agnostic classifier, for example by asserting that each training domain's classifier is the sum of a domain-specific and domain-agnostic weight vector \cite{ECCV12_Khosla}. The resulting domain-agnostic weight vector can then be extracted and applied to held out domains. Our approach lies in this latter category. However, prior work in this area has dealt with shallow, linear models only. We show how to extend this intuition to end-to-end learning in CNNs, while limiting the resulting parameter growth, and making the sharing structure richer than an unweighted sum.

There has been more extensive work on CNN models for domain \emph{adaptation}, with methods developed for encouraging CNN layers to learn transferable features \cite{ganin2015udaBackprop,long2015deepAdaptationNetworks}. However, these studies have typically not addressed our domain \emph{generalization} setting. Moreover, as analysis has shown that the transferability of different layers in CNNs varies significantly \cite{yosinski2014howTransferable}, these studies have had carefully hand designed the CNN sharing structure to address their particular DA problems. In our benchmark, this is harder, as the gaps between our more diverse domains are unknown and likely to be more variable. However, our low-rank modeling approach provides the benefit of automatically estimating both the per-domain and per-layer sharing strength.

Domain Generalization is also related to \emph{learning to learn}. Learning to learn methods aim to learn not just specific concepts or skills, but learning algorithms or problem agnostic biases that improve generalization \cite{metanetworks2017,ravi2016optimization,finn2017model}.
Similarly  DG is to extract common knowledge from source domains that applies to unseen target domains.
Thus our method can be seen as a simple learning to learn method for the DG setting. Different from few-shot learning \cite{finn2017model,ravi2016optimization}, DG is a zero-shot problem as performance is immediately evaluated on the target domain with no further learning. 

\vspace{0.1cm}\noindent\textbf{Neural Network Methods}\quad
Our DG method is related to parameterized neural networks \cite{bertinetto2016feedForwardOneShot,sigaud2015gatedInventory}, in that the parameters are set based on external metadata. In our case, based on a description of the current domain, rather than an instance \cite{bertinetto2016feedForwardOneShot}, or additional sensor \cite{sigaud2015gatedInventory}. It is also related to low-rank neural network models, typically used to compress \cite{kim2016compressionCNN} and speed up \cite{lebedev2015cpDecompCNN} CNNs, and have very recently been explored for cross-category CNN knowledge transfer \cite{yang2017deepMTRL}. In our case we exploit this idea both for compression -- but across rather than within domains \cite{kim2016compressionCNN}, as well as for cross-domain (rather than cross-category \cite{yang2017deepMTRL}) knowledge sharing. Different domains can share parameters via common latent factors. \cite{bousmalis2016domain} also addresses the DG setting, but learns  shared parameters based on image reconstruction, whereas ours is learned via paramaterizing each domain's CNN.
As a parameterized neural network, our approach also differs from all those other low-rank methods \cite{kim2016compressionCNN,lebedev2015cpDecompCNN,yang2017deepMTRL}, which have a fixed parameterization.

\subsection{Benchmarks and Datasets} 
\noindent\textbf{DG Benchmarks}\quad
The most popular DG benchmarks are: `Office' \cite{saenko2010domainAdapt} (containing Amazon/Webcam/DSLR images), later extended to include a fourth Caltech 101 domain \cite{gong2012geodesicFlowDA} (OfficeCaltech) and Pascal 2007, LabelMe, Caltech, SUN09 (VLCS)  \cite{torralba2011dataset_bias,ECCV12_Khosla}. The domains within Office relate to different camera types, and the others are created by the biases of different data collection procedures \cite{torralba2011dataset_bias}. Despite the famous analysis of dataset bias \cite{torralba2011dataset_bias} that motivated the creation of the VLCS benchmark, it was later shown that the domain shift is much smaller with recent deep features \cite{donahue2014decaf}. Thus recent DG studies have used deep features \cite{ghifary2015domain}, to obtain better results. Nevertheless, we show that a very simple baseline of fine-tuning deep features on multiple source domains performs comparably or better than prior DG methods. This motivates our design of a CNN-based DG method, as well as our new dataset (Fig~\ref{fig:datasetSummary}) which  has greater domain shift than the prior benchmarks.  Our dataset draws on non-photorealistic and abstract visual domains which provide a better motivated example of the sort of relatively sparse data domain where DG would be of practical value.

\noindent\textbf{Non-photorealistic Image Analysis}\quad
Non-photorealistic image analysis is a growing subfield of computer vision that extends the conventional photo-only setting of vision research to include other visual depictions (often more abstract) such as paintings and sketches. Typical tasks include instance-level matching between sketch-photo \cite{yu2016sketchShoe,sangkloy2016sketchy}, and art-photo domains \cite{crowley2015facePainting}, and transferring of object recognizers trained on photos to detect objects in art \cite{crowley2016artDetection,wu2014matchAcrossStyle}. Most prior work focuses on two domains (such as photo and painting \cite{crowley2016artDetection,wu2014matchAcrossStyle}, or photo and sketch \cite{yu2016sketchShoe,sangkloy2016sketchy}). Studies have investigated simple `blind' transfer between domains \cite{crowley2016artDetection}, learning cross-domain projections \cite{yu2016sketchShoe,crowley2015facePainting}, or engineering structured models for matching \cite{wu2014matchAcrossStyle}. Thus, in contrast to our DG setting, prior non-photorealistic analyses fall into either cross-domain instance matching, or domain adaptation settings. To create our benchmark, we aggregate multiple domains including paintings, cartoons and sketches, and define a comprehensive domain-generalization benchmark covering a wide spectrum of visual abstraction based upon these. Thus in contrast to prior DG benchmarks, our domain-shifts are bigger and more challenging. 

\section{Methodology}

Assume we observe $S$ domains, and the $i$th domain contains $N_i$ labeled instances  $\{(x_j^{(i)},y_j^{(i)})\}^{N_i}_{j=1}$ where $x_j^{(i)}$ is the input data (e.g., an image) for which we assume they are of the same size among all domains (e.g., all images are cropped into the same size), and $y_j^{(i)}\in \{1\dots C\}$ is the class label. We assume the label space is consistent across domains. The objective of DG is to learn a domain agnostic model which can be applied to unseen domains in the future. In contrast to domain adaptation, we can not access the labeled or unlabeled examples from those domains to which the model is eventually applied. So the model is supposed to extract the domain agnostic knowledge within the observed domains. In the training stage, we will minimize the empirical error for all observed domains,
\begin{equation}
\underset{\Theta_1,\Theta_2,\dots,\Theta_S}{\operatorname{argmin}} \frac{1}{S} \sum^{S}_{i=1}\frac{1}{N_i}\sum^{N_i}_{j=1} \ell(\hat{y}^{(i)}_j,y^{(i)}_j)
\end{equation}
\noindent where $\ell$ is the loss function that measures the error between the predicted label $\hat{y}$ and the true label $y$, and  prediction is carried out by a function $\hat{y}^{(i)}_j = f(x^{(i)}_j | \Theta_i)$ parameterized by $\Theta_i$. A straightforward approach to finding a domain agnostic model is to assume $\Theta_* = \Theta_1 = \Theta_2 = \cdots = \Theta_S$, i.e., there exists a universal model $\Theta_*$. Doing so we literally ignore the domain difference. Alternatively, Undo-Bias \cite{ECCV12_Khosla} considers linear models, and assumes that the parameter (a $D$-dimensional vector when $x\in\mathbb{R}^D$) for the $i$th domain is in the form $\Theta^{(i)}=\Theta^{(0)}+\Delta^{(i)}$, where $\Theta^{(0)}$ can be seen as a domain agnostic model that benefits all domains, and $\Delta^{(i)}$ is a domain specific bias term. Conceptually, $\Theta^{(0)}$ can also serve as the classifier  for any unseen domains. \cite{ECCV12_Khosla} showed that (for linear models) $\Theta^{(0)}$ is better than the universal model $\Theta_*$ trained by $\underset{\Theta_*}{\operatorname{argmin}} \frac{1}{S} \sum^{S}_{i=1}\frac{1}{N_i}\sum^{N_i}_{j=1} \ell(\Theta_*^T x^{(i)}_j,y^{(i)}_j)$ in terms of testing performance on unseen domains. However we show that for deep networks, a universal model $f(x|\Theta_*)$ is a strong baseline that requires improved methodology to beat.

\subsection{Parameterized Neural Network for DG}

To extend the idea of Undo-Bias \cite{ECCV12_Khosla} into the neural network context, it is more convenient to think $\Theta^{(i)}$ is \emph{generated} from a function $g(z^{(i)} | \Theta)$ parameterized by $\Theta$. Here $z^{(i)}$ is a binary vector encoding of the $i$th domain with two properties: (i) it is of length $S+1$ where $S$ is the number of observed domains; (ii) it always has only two units activated (being one): the $i$th unit active for the $i$th domain and the last unit active for all domains. Formally, the objective function becomes,

\begin{equation}
\underset{\Theta}{\operatorname{argmin}} \frac{1}{S} \sum^{S}_{i=1}\frac{1}{N_i}\sum^{N_i}_{j=1} \ell(\hat{y}^{(i)}_j,y^{(i)}_j)
\end{equation}

\noindent where $\hat{y}^{(i)}_j = f(x^{(i)}_j | \Theta_i) = f(x^{(i)}_j | g(z^{(i)} | \Theta))$.

To reproduce Undo-Bias \cite{ECCV12_Khosla}, we can stack all parameters in a column-wise fashion to form $\Theta$, i.e., $\Theta = [\Delta^{(1)}, \Delta^{(2)}, \dots, \Delta^{(S)}, \Theta^{(0)}]$, and choose the $g(\cdot)$ function to be linear mapping: $g(z^{(i)} | \Theta) = \Theta z^{(i)}$.

\vspace{0.1cm}\noindent\textbf{From linear to multi-linear}\quad The method as described so far generates the model parameter in the form of \emph{vector} thus it is only suitable for single-out setting (univariate regression or binary classification). To generate higher order parameters, we use a multi-linear model, where $\Theta$ is (3rd order or higher) tensor. E.g., to generate a weighting matrix for a fully-connected layer in neural network, we can use

\begin{equation}
W^{(i)}_{\text{FC}}=g(z^{(i)} | \mathcal{W}) = \mathcal{W}\times_3 z^{(i)}\label{eq:fc-decom}
\end{equation}

\noindent Here $\times_3$ is the inner product between tensor and vector along tensor's $3$rd axis. For example if $W$ is the weight matrix of size $H\times C$ (i.e., the number of input neurons is $H$ and the number of output neurons is $C$) then $\mathcal{W}$ is a $H\times C\times (S+1)$ tensor.

If we need to generate the parameter for a convolutional layer of size $D_1\times D_2 \times F_1 \times F_2$ ($\text{Height} \times \text{Width} \times \text{Depth} \times \text{Filter Number}$), then we use:
\begin{equation}
\mathcal{W}^{(i)}_{\text{CONV}}=g(z^{(i)} | \mathcal{W}) = \mathcal{W}\times_5 z^{(i)}
\end{equation}
\noindent where $\mathcal{W}$ is a $5$th order tensor of size $D_1\times D_2 \times F_1 \times F_2\times (S+1)$.

\vspace{0.1cm}\noindent\textbf{Domain generalization}\quad
Using one such parameter generating function per layer, we can dynamically generate the weights at \emph{every} layer of a CNN based on the encoded vector of every domain. In this approach, knowledge sharing is realized through the last (bias) bit in the encoding of $z$. I.e., every weight tensor for a given domain is the sum of a domain specific tensor and a (shared) domain agnostic tensor. For generalization to an unseen domain, we apply the one-hot, bias-only, vector $z_*=[0,0,\dots,0,1]$ to synthesize a domain agnostic CNN. 

\subsection{Low rank parameterized CNNs}

The method as described so far has two limitations: (i) the required parameters to learn now grow linearly in the number of domains (which we eventually hope to be large to achieve good DG), and (ii) the sharing structure is very prescribed: every parameter is an equally weighted sum of its domain agnostic and domain-specific bias partners. 

To alleviate these two issues, we place a structural constraint on $\mathcal{W}$. Motivated by the well-known Tucker decomposition \cite{Tuck1966c}, we assume that the $M$-order tensor $\mathcal{W}$ is synthesized as:
\begin{equation}
\mathcal{W} = \mathcal{G} \times_1 U_1 \dots \times_M U_M
\end{equation}
\noindent where $\mathcal{G}$ is a $K_1\times \dots K_M$  sized low-rank core tensor, and $U_m$ are $K_m \times D_m$ matrices (note that $D_M = S+1$). By controlling the ranks $K_1\dots K_M$ we can effectively reduce the number of parameters to learn. By learning $\{\mathcal{G},U_1\dots U_M\}$ instead of $\mathcal{W}$, the number of parameters is reduced from $(D_1\times\dots\times  D_{M-1} \times (S+1))$ to $(K_1\times\dots K_M) + \sum_{m=1}^{M-1} D_m\times K_m + K_M\times(S+1)$. Besides, $U_M$ produces a $K_M$-dimensional dense vector that guides how to linearly combine the shared factors, which is much more informative than the original case of equally weighted sum.

Given a tensor $\mathcal{W}$ the Tucker problem can be solved via high-order singular value decomposition (HO-SVD) \cite{Lathauwer2000}. 
\begin{equation}
\mathcal{G} = \mathcal{W}\times_1 U^T_1 \dots \times_M U^T_M\label{eq:svdTucker}
\end{equation}
where $U_n$ is the $U$ matrix from the SVD of the the mode-$n$ flattening of $\mathcal{W}$. However, note that aside from (optionally) performing this once for initialization, we do \emph{not} perform this costly HO-SVD operation during learning. 

\vspace{0.1cm}\noindent\textbf{Inference and Learning}\quad To make predictions for a particular domain, we synthesize a concrete CNN by multiplying out the parameters $\{\mathcal{G},U_1,\dots,U_M\}$ after that doing an inner product with the corresponding domain's $z$. This CNN can then be used to classify an input instance $x$. Since our method does not introduce any non-differentiable functions, we can use standard back-propagation to learn $\{\mathcal{G},U_1,\dots,U_M\}$ for every layer.

For our model there are hyperparameters -- Tucker rank $[K_1\dots K_M]$ -- that can potentially be set at each layer. We sidestep the need to set all of these, by using the strategy of decomposing the stack of (ImageNet pre-trained) single domain models plus one agnostic domain model through Tucker decomposition, and then applying a reconstruction error threshold of $\epsilon=10\%$ for the HO-SVD in Eq~\ref{eq:svdTucker}. This effectively determines all rank values via one `sharing strength' hyperparameter $\epsilon$.

\section{Experiments}
\subsection{New Domain Generalization Dataset: PACS}\label{sub:ourDataset}
Our \textcolor{black}{PACS} DG dataset is created by intersecting the classes found in Caltech256 (Photo), Sketchy (Photo, Sketch) \cite{sangkloy2016sketchy}, TU-Berlin (Sketch) \cite{eitz2012howToSketch} and Google Images (Art painting, Cartoon, Photo). Our dataset and code, together with latest results using alternative state-of-the-art base networks, can be found at: \url{http://sketchx.eecs.qmul.ac.uk/}.

\noindent\textbf{PACS:}\quad Our new benchmark includes 4 domains (Photo, Sketch, Cartoon, Painting), and 7 common categories `dog', `elephant', `giraffe'. `guitar', `horse', `house', `person'. The total number of images is 9991. 

\begin{figure*}[t]
\centering
\begin{subfigure}{0.14\textwidth}
\centering
\vspace{1cm}
\includegraphics[width=1.0\linewidth]{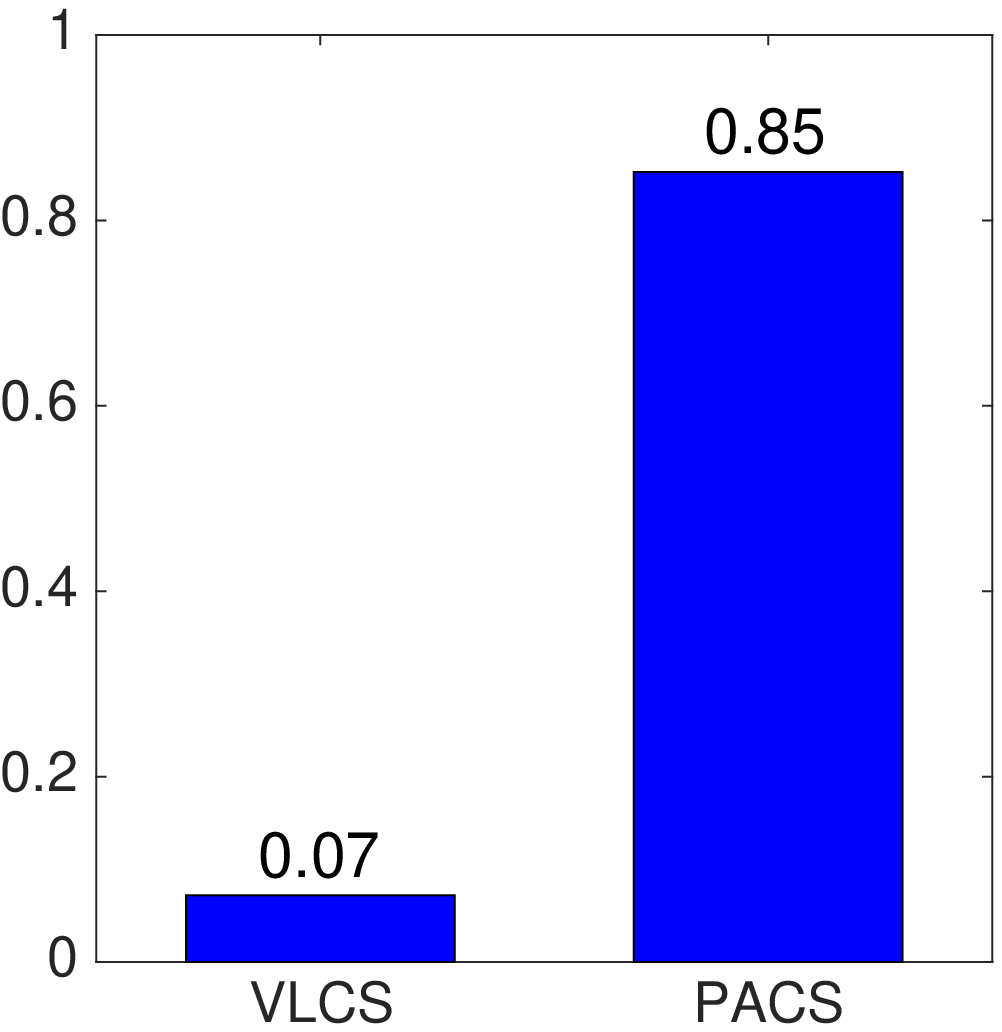}
\caption{Average KLD between domains.}
\end{subfigure}
\hspace{0.6cm}
\begin{subfigure}{0.32\textwidth}
\centering
\includegraphics[width=1.0\linewidth]{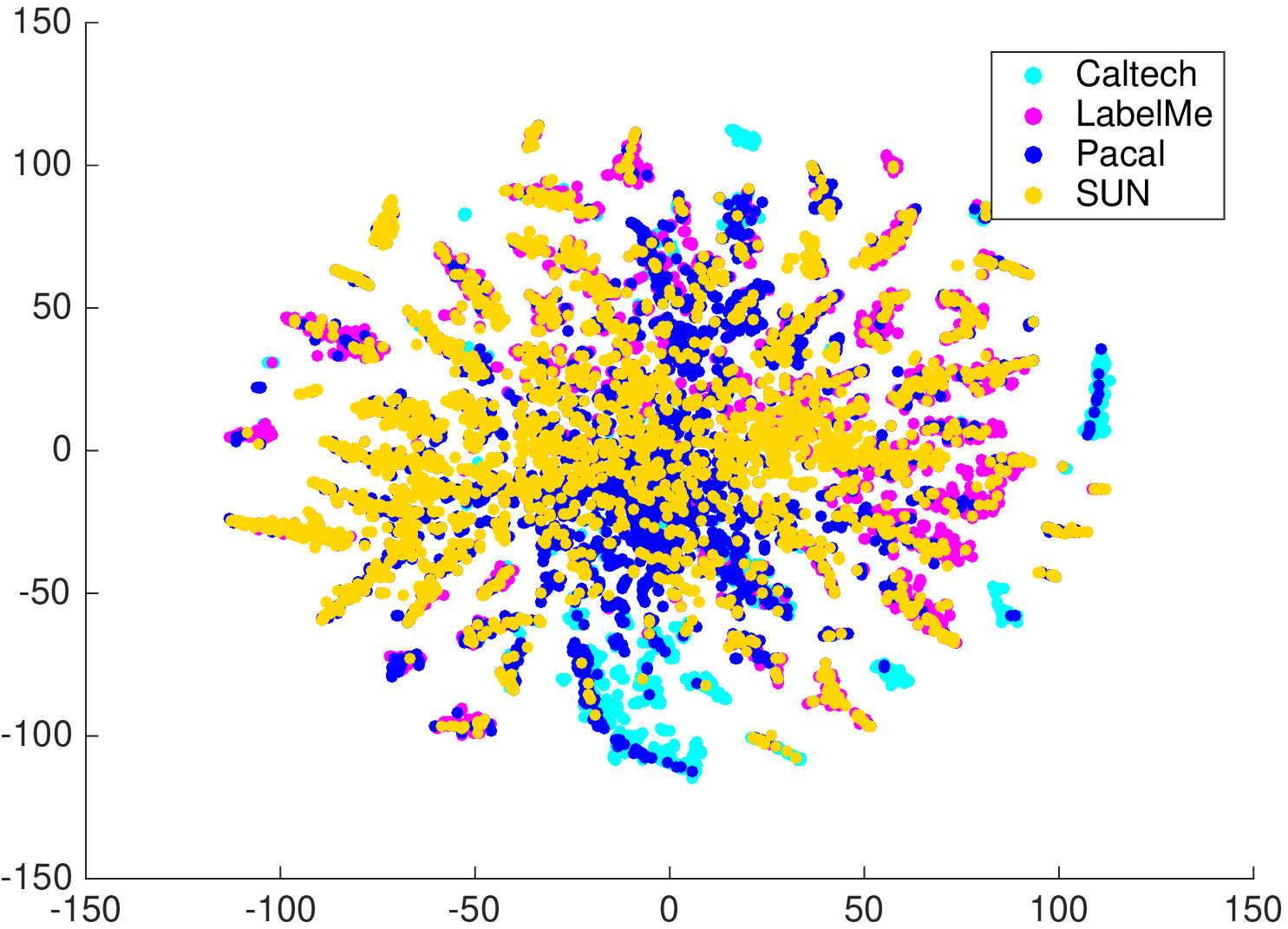}
\caption{The feature distribution of VLCS}
\end{subfigure}
\hspace{-0.2cm}
\begin{subfigure}{0.32\textwidth}
\centering
\includegraphics[width=1.\linewidth]{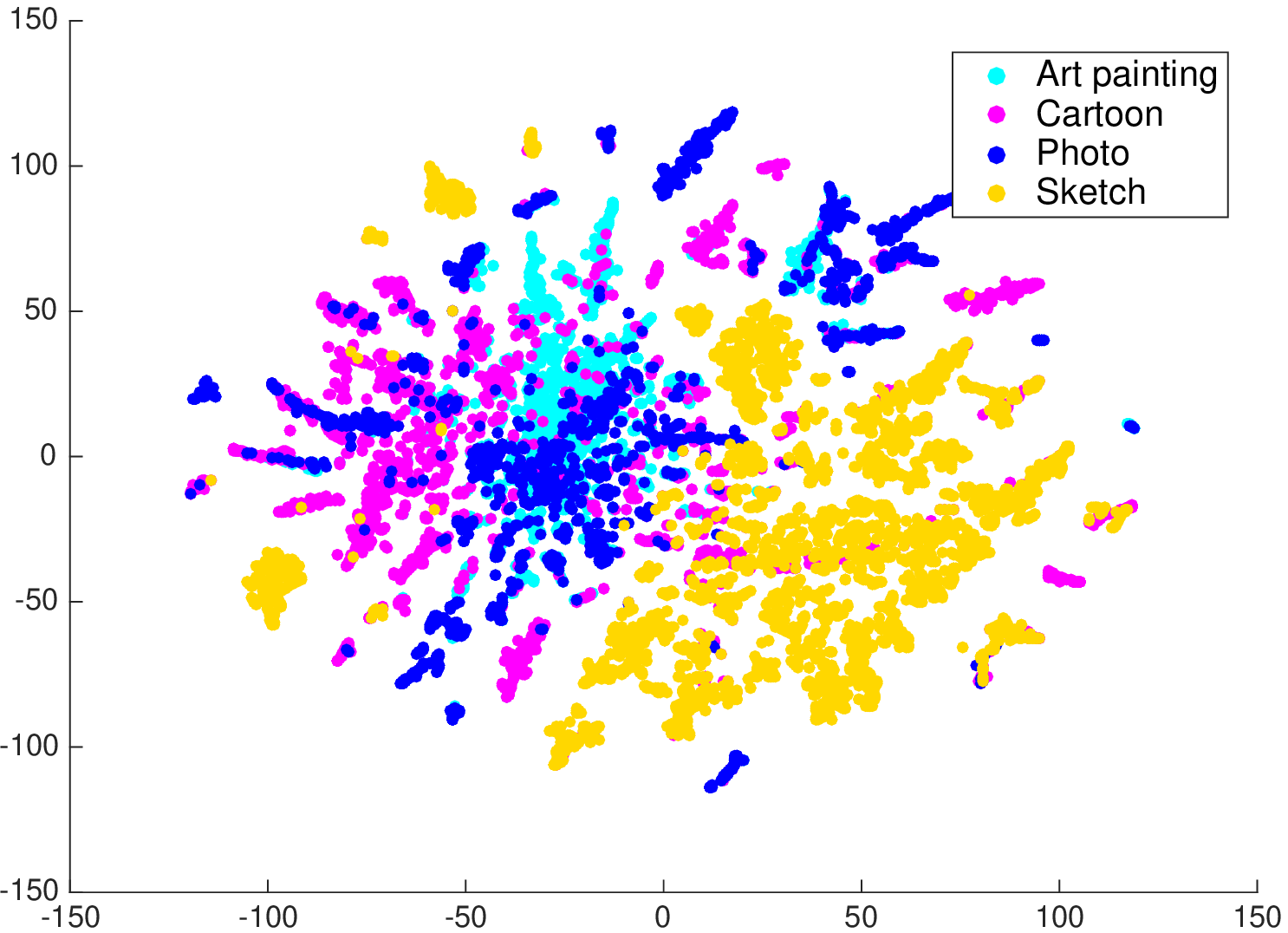}
\caption{The feature distribution of PACS}
\end{subfigure}
\caption{Evaluation of domain shift in different domain generalization benchmarks.}\label{fig:domain-bias}
\end{figure*}

\subsection{Characterizing Benchmarks' Domain  Shifts}\label{sub:datasetAnalysis}
We first perform a preliminary analysis to contrast the domain shift within our PACS dataset to that of prior popular datasets such as VLCS. We make this contrast from both a feature space and a classifier performance perspective.\\
\noindent\textbf{Feature Space Analysis}\quad
Given the DG setting of training on source domains and applying to held out test domain(s), we measure the shift between source and target domains based on the Kullback-Leibler divergence as: $D_{shift}(D^{s}, D^{t})= \frac{1}{m\times n} \sum\limits_i^n\sum\limits_j^m \lambda_{i} KLD(D_{i}^{s}||D_{j}^{t})$,
 where $n$ and $m$ are the number of source and target domains, and $\lambda_{i}$ weights the $i$ th source domain, to account for data imbalance. To encode each domain as a probability, \textcolor{black}{we calculate the mean DECAF$_7$ representation over instances and then apply softmax normalization}. 

\vspace{0.1cm}\noindent\textbf{Classifier Performance Analysis}\quad We also compare the datasets by the margin between multiclass classification accuracy of within-domain learning, and a simple cross-domain baseline of training a CNN on all the source domains before testing on the held out target domain (as we shall see later, this baseline is very competitive). Assuming within-domain learning performance is an upper bound, then this difference indicates the space which a DG method has to make a contribution, and hence roughly reflects size of the domain-shift/difficulty of the DG task. 

\begin{figure*}[t]
\centering
\begin{subfigure}{0.70\textwidth}
\includegraphics[width=1.0\columnwidth]{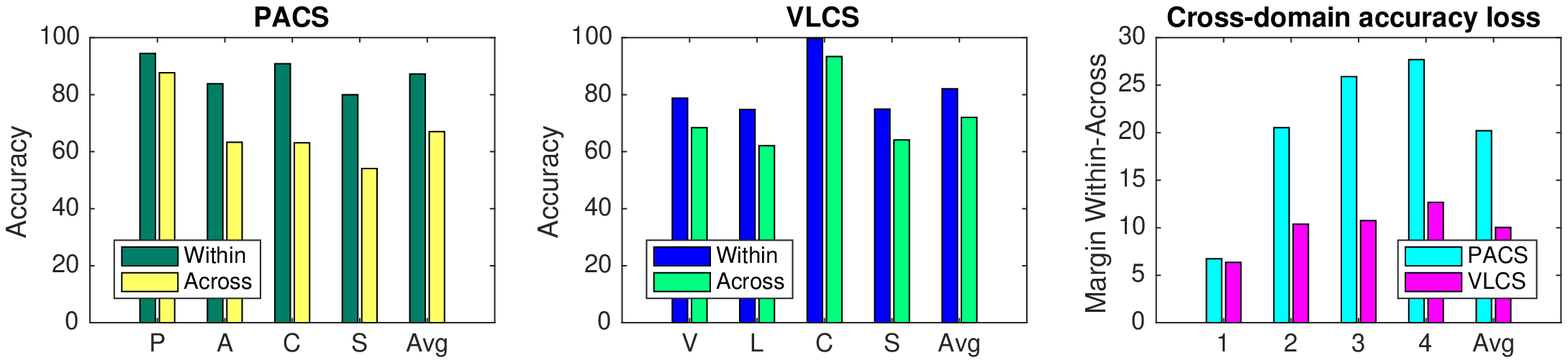}
\caption{Measuring domain-shift by within versus across domain accuracy. Left: Our PACS, Middle: VLCS. Right: Distribution of margins between within and across domain accuracy.}\label{fig:crossDomainAcc}
\end{subfigure}
\hspace{0.3cm}
\begin{subfigure}{0.26\textwidth}
\includegraphics[width=1.0\columnwidth]{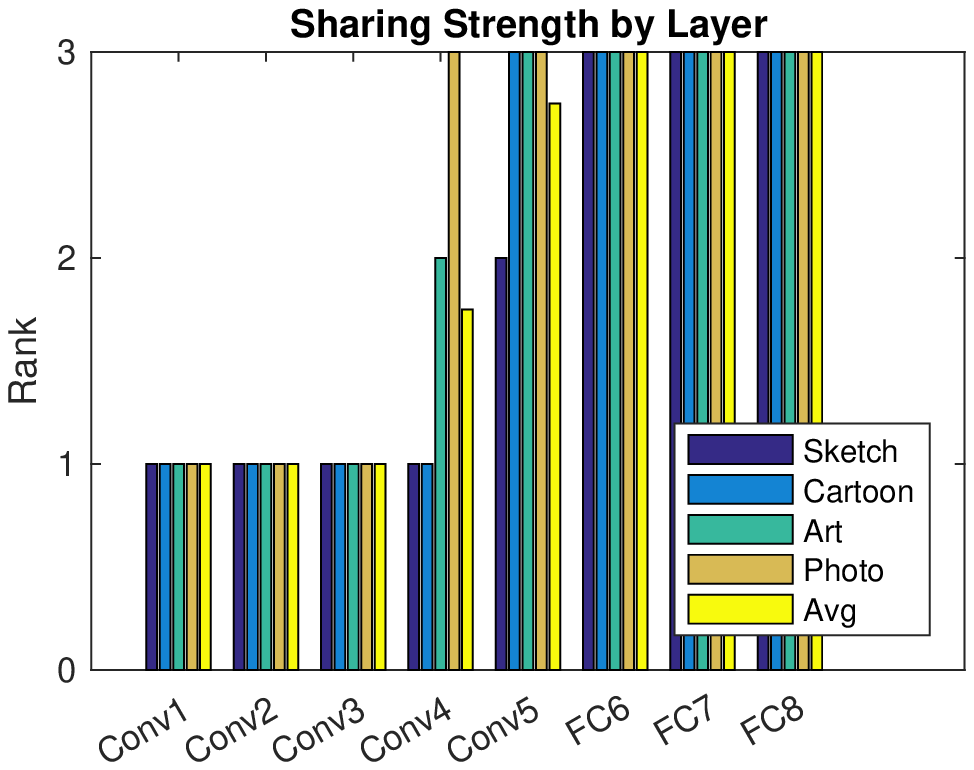}
\caption{Per-layer rank (inverse sharing strength) after learning for different held out domains.}
\label{fig:sharing-strength}
\end{subfigure}
\caption{Cross-domain similarity (a) and learned sharing strength by layer (b).}
\end{figure*}



\vspace{0.2cm}\noindent\textbf{Results}\quad
Fig.~\ref{fig:domain-bias}(a) shows the average domain-shift in terms of KLD across all choices of held out domain in our new PACS benchmark, compared with the VLCS benchmark \cite{torralba2011dataset_bias}. Clearly the domain shift is significantly higher in our new benchmark, as is visually intuitive from the illustrative examples in Fig.~\ref{fig:datasetSummary}. To provide a qualitative summarization, we also show the distribution of features in our PACS compared to VLCS in Fig.~\ref{fig:domain-bias}(b,c) as visualized by a 2 dimensional t-SNE \cite{maaten2008visualizing} plot, where the features are categorized and colored by their associated domain. From this result, we can see that the VLCS data are generally hard to separate by domain, while our PACS data are much more separated by domain. This illustrates the greater degree of shift between the domains in PACS over VLCS. 


We next explore the domain shifts from a model-, rather than feature-centric perspective. Fig.~\ref{fig:crossDomainAcc} summarizes the within-domain and across-domain performance for each domain within PACS and VLCS benchmarks. The average drop in performance due to cross-domain transfer is $20.2\%$ for PACS versus $10.0\%$ for VLCS. This shows that the scope for contribution of DG/DA in our PACS is double that of VLCS, and illustrates the greater relevance and challenge of the PACS benchmark.

\subsection{Domain Generalization Experiments}
\subsubsection{Datasets and Settings}
We evaluate our proposed method on two datasets:  VLCS, and our proposed PACS dataset. \textbf{VLCS} \cite{torralba2011dataset_bias} aggregates photos from Caltech, LabelMe, Pascal VOC 2007 and SUN09. It provides a 5-way multiclass benchmark on the five common classes: 'bird','car','chair','dog' and 'person'. 
Our \textbf{PACS}  (described in Sec.~\ref{sub:ourDataset}) with 7 classes
from Photo, Sketch, Cartoon, Painting domains. 
All results are evaluated by multi-class accuracy, following \cite{ghifary2015domain}. We explore features including \textbf{Classic} SIFT features (for direct comparison with earlier work), \textbf{DECAF} pre-extracted deep features following \cite{ghifary2015domain}, and \textbf{E2E} end-to-end CNN learning.

\vspace{0.1cm}\noindent\textbf{Settings:}\quad For our method in E2E configuration, we use the ImageNet pre-trained AlexNet CNN, fine-tuned with multi-domain learning on the training domains. On VLCS, we follow the train-test split strategy from \cite{ghifary2015domain}. 
Our initial learning rate is 5e-5 and batch size is 64 for each training domain. We use the best performed model on validation to do the test after tuning the model for 25$k$ iterations.  On PACS, we split the images from training domains to 9 (train)~:~1 (val) and test on the whole held-out domain. 
Recall that our model uses a 2-hot encoding of $z$ to parameterize the CNN. The domain-specific vs agnostic `prior' can be set by varying the ratio $\rho$ of the elements in the 2-hot coding. For training we use $\rho=0.3$, so  $z=\{[0, 0, 0.3, 1]$, $[0, 0.3, 0, 1],...\}$. For DG testing  we use $z=[0,0,0,1]$.
\begin{table*}[t]
\small
\begin{center}
\scalebox{0.86}{
\begin{tabular}{ ccccccccccc }
 \hline
 \multirow{2}{*}{Unseen domain}  &  \multicolumn{2}{c}{Bird} & \multicolumn{2}{c}{Car}& \multicolumn{2}{c}{Chair}& \multicolumn{2}{c}{Dog}& \multicolumn{2}{c}{Person} \\
 & Undo bias & Ours-MLP & Undo bias& Ours-MLP & Undo bias& Ours-MLP & Undo bias & Ours-MLP & Undo bias & Ours-MLP \\
 \hline
 Caltech  & \textbf{12.08} & 10.89 & \textbf{63.80} & 61.29 & 7.54 & \textbf{11.26} & \textbf{5.24} & 3.90 & \textbf{50.81} & 48.48 \\ 

 LabelMe & \textbf{33.08} & 28.35 & 69.22 & \textbf{74.07} & \textbf{5.34} & 3.68 & 1.66 & \textbf{2.06} & 64.85 & \textbf{67.00} \\

 Pascal & \textbf{15.42} & 13.63 & 37.49 & \textbf{42.81} & 30.05 & \textbf{32.71} & 14.97 & \textbf{15.93} & 58.47 & \textbf{63.61} \\
 
 Sun & 0.59 & \textbf{2.01} & 70.62 & \textbf{71.32} & 37.44 & \textbf{37.50} & 1.12 & \textbf{1.89} & 42.20 & \textbf{42.71} \\
 \hline
 Mean AP \% & \textbf{15.29} & 13.72 & 60.28 & \textbf{62.37} & 20.09 & \textbf{21.29} & 5.75 & \textbf{5.94} & 54.08 & \textbf{55.45}  \\
 \hline
\end{tabular}
}
\caption{Comparison against Undo-Bias \cite{ECCV12_Khosla} on the VLCS benchmark using classic SIFT-BOW features, and our shallow model Ours-MLP. Average precision (\%) and mean average precision (\%) of binary 1-v-all classification in unseen domains.}
\label{tab:binary+shallow}
\end{center}
\end{table*}
\begin{table*}[t]
\centering
\scalebox{0.68}{
\begin{tabular}{cccccccccc|cc}
\hline
\multirow{2}{*}{Unseen domain} & \multicolumn{9}{c|}{Image $\mapsto$ Deep Feature $\mapsto$ Classifier} & \multicolumn{2}{|c}{Image $\mapsto$ E2E}                        \\
             &SVM &1HNN &Undo-Bias\cite{ECCV12_Khosla} &uDICA\cite{muandet2013domainGen} &UML\cite{fang2013unbiased} &LRE-SVM\cite{xu2014exploiting}&  MTAE+1HNN\cite{ghifary2015domain} & D-MTAE+1HNN\cite{ghifary2015domain} & Ours-MLP & Deep-All & Ours-Full \\
              \hline
Caltech      &77.67&86.67 &87.50& 61.70 &91.13&88.11& 90.71 & 89.05    & 92.43 & 93.40  &   \textbf{93.63}  \\
LabelMe      &52.49&58.20 &58.09& 46.67 &58.50&59.74& 59.24 & 60.13    & 58.74 & 62.11  &   \textbf{63.49}   \\
Pascal       &58.86&59.10 &54.29& 44.41 &56.26&60.58& 61.09 & 63.90    & 65.58 & 68.41  &   \textbf{69.99}   \\
Sun          &49.09&57.86 &54.21& 38.56 &58.49&54.88& 60.20 & 61.33    & 61.85 & \textbf{64.16}  &   61.32   \\ \hline
Ave.\%       &59.93&65.46 &63.52& 47.83 &65.85&65.83& 67.81 & 68.60    & 69.65 & 72.02  &   \textbf{72.11}   \\\hline 
\end{tabular}
}
\caption{Comparison of features and state of the art on the VLCS benchmark. Multi-class accuracy (\%).}
\label{tab:deepCLPS}
\end{table*}

\vspace{0.1cm}\noindent\textbf{Baselines:}\quad We evaluate our contributions by comparison with number of alternatives including variants designed to reveal insights, and state of the art competitors:\\
\textbf{Ours-MLP:} Our DG method applied to a 1 hidden layer multi-layer perception. For use with pre-extracted features.\\
\textbf{Ours-Full:} Our full low-rank parameterized CNN trained end-to-end on images.
\textbf{SVM:} Linear SVM, applied on the aggregation of data from all source domains. 
\textbf{Deep-All:} Pretrained Alexnet CNN \cite{krizhevsky2012imagenet},  fine-tuned on the aggregation of all source domains. 
\textbf{Undo-Bias:} Modifies traditional SVM to include a domain-specific and global weight vector which can be extracted for DG \cite{ECCV12_Khosla}. The original Undo-Bias is a binary classifier (BC). We also implement a multi-class (MC) generalization. 
\textbf{uDICA:} A kernel based method learning a subspace to minimize the dissimilarity between domains \cite{muandet2013domainGen}\footnote{Like \cite{ghifary2015domain}, we found sDICA to be worse than uDICA, so excluded it.}.
\textbf{UML}:  Structural metric learning algorithm learn a low-bias distance metric for classification tasks \cite{fang2013unbiased}. 
\textbf{LRE-SVM:}  Exploits latent domains, and a nuclear-norm based regularizer on the likelihood matrix of exemplar-SVM \cite{xu2014exploiting}.
\textbf{1HNN}: 1 hidden layer neural network.
\textbf{MTAE-1HNN}: 1HNN with multi-task auto encoder \cite{ghifary2015domain}. \textbf{D-MTAE-1HNN}: 1HNN with de-noising multi-task auto encoder \cite{ghifary2015domain}.
\textbf{DSN}: The domain separation network learns specific and shared models for the source and target domains \cite{bousmalis2016domain}. We re-purpose the original DSN from the domain adaptation to the DG task. Note that DSN is already shown to outperform the related \cite{ganin2015udaBackprop}.


\subsubsection{VLCS Benchmark}
\noindent\textbf{Classic Benchmark - Binary Classification with Shallow Features}\quad
Since our approach to extracting a domain invariant model is related to the intuition in Undo Bias \cite{ECCV12_Khosla}, we first evaluate our methodology by performing a direct comparison against Undo Bias. We use the same 5376 dimensional VLCS SIFT-BOW features\footnote{\url{http://undoingbias.csail.mit.edu/}} from \cite{ECCV12_Khosla}, and compare Our-MLP using one RELU hidden layer with 4096 neurons. For direct comparison, we apply Our-MLP in a 1-vs-All manner as per Undo-Bias. The results in Table~\ref{tab:binary+shallow} show that without exploiting the benefit of end-to-end learning, our approach still performs favorably compared to Undo Bias. This is due to (i) our low-rank modeling of domain-specific and domain-agnostic knowledge, and (ii) the generalization of doing so in a multi-layer network.


\vspace{0.2cm}\noindent\textbf{Multi-class recognition with Deep Learning}\quad
In this experiment we continue to analyze the VLCS benchmark, but from a multiclass classification perspective. We compare existing DG methods (Undo-Bias \cite{ECCV12_Khosla}, UML \cite{fang2013unbiased}, LRE-SVM \cite{xu2014exploiting}, uDICA \cite{muandet2013domainGen}, MTAE+1HNN \cite{ghifary2015domain}, D-MTAE+1HNN \cite{ghifary2015domain}) against baselines (1HNN, SVM, Deep) and our methods Ours-MLP/Ours-Full. For the other methods besides Deep-All and Ours-Full, we follow \cite{ghifary2015domain} and use pre-extracted DECAF$_6$ features\footnote{\url{http://www.cs.dartmouth.edu/~chenfang/proj_page/FXR_iccv13/index.php}} \cite{donahue2014decaf}. For Deep and Ours-Full, we fine-tune the CNN on the source domains.

From the results in Table~\ref{tab:deepCLPS}, we make the following observations: (i) Given the fixed DECAF$_6$ feature, most prior DG methods improve on vanilla SVM, and D-MTAE \cite{ghifary2015domain} is the best of these. (ii) Ours-MLP outperforms 1HNN, which uses the same type of architecture 
and the same feature. This margin is due to our low-rank domain-generalization approach. (iii) The very simple baseline of fine-tuning a deep model on the aggregation of source domains (Deep-All) performs surprisingly well and actually outperforms all the prior DG methods. (iii) Ours-Full outperforms Deep-All slightly. This small margin is understandable. Our model does have more parameters to learn than Deep-All, despite the low rank; and the cost of doing this is not justified by the relatively small domain gap between the VLCS datasets.

\begin{figure*}[!t]
\centering
\includegraphics[width=0.95\linewidth]{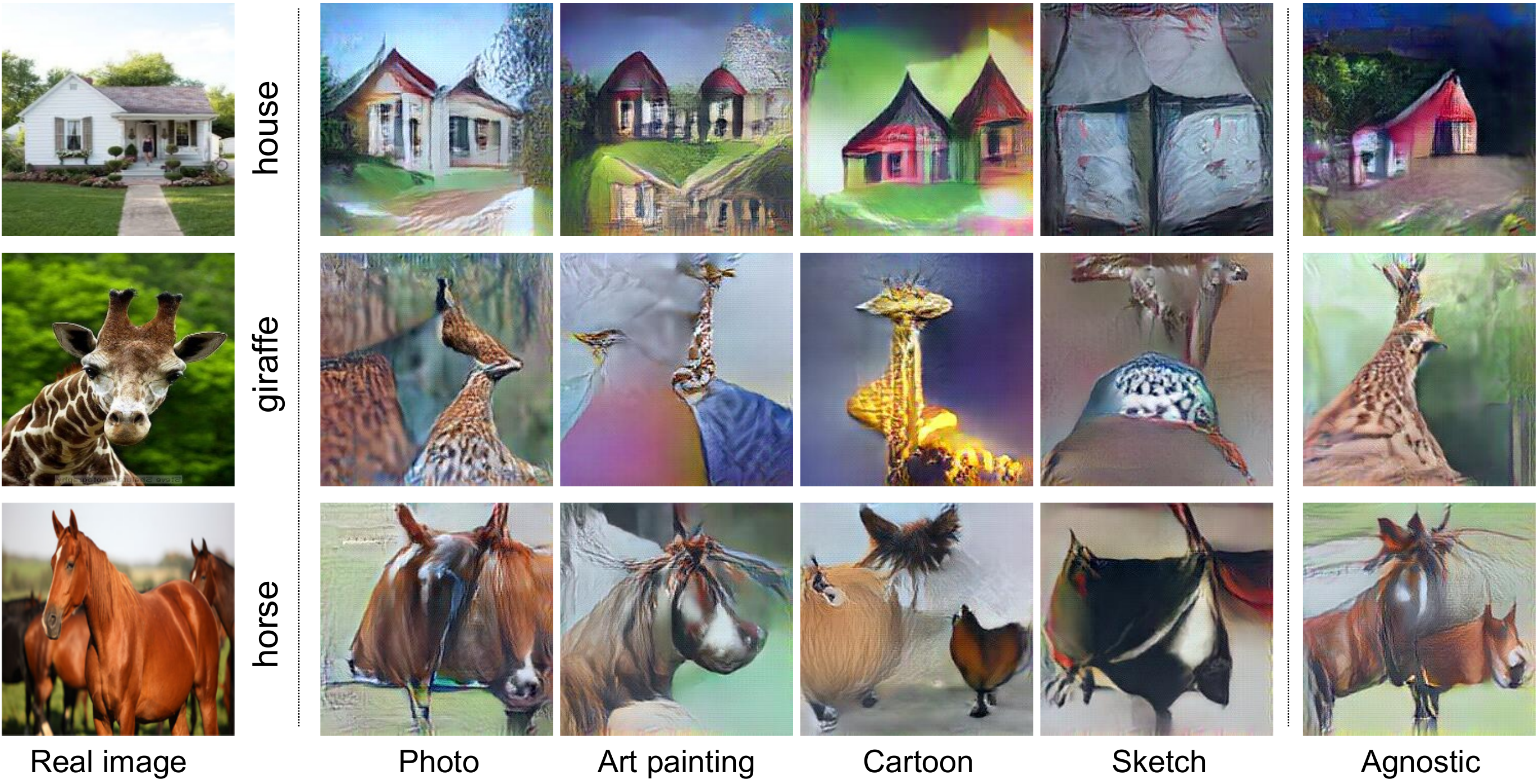}
\caption{Visualization of the preferred images of output neurons `horse', `giraffe' and `house' in the domains of the PACS dataset. Left: real images. Middle: synthesized images for PACS domains. Right: synthesized images for agnostic domain.}
\label{fig:visualization4domain}
\end{figure*}


\begin{table*}[t]
\centering
\scalebox{0.85}{
\begin{tabular}{ccccccc|ccc}
\hline
\multirow{2}{*}{Unseen domain} & \multicolumn{6}{c|}{Image $\mapsto$ Deep Feature $\mapsto$ Classifier} & \multicolumn{3}{|c}{Image $\mapsto$ E2E}    \\
              & SVM & 1HNN   & uDICA \cite{muandet2013domainGen} & LRE-SVM \cite{xu2014exploiting} & D-MTAE+1HNN \cite{ghifary2015domain}  & Ours-MLP & Deep-All &DSN \cite{bousmalis2016domain} & Ours-Full \\
              \hline
Art painting & 55.39           & 59.10    & \textbf{64.57} &59.74   & 60.27          	&  61.40  & 63.30   &61.13				  & 62.86               \\
Cartoon       & 52.86           & 57.89   & 64.54 & 52.81  & 58.65          			&  57.16  & 63.13   &66.54			       & \textbf{66.97}       \\
Photo         & 82.83           & 89.86   & \textbf{91.78} & 85.53  & 91.12 			&  89.68  & 87.70   &83.25			       & 89.50       \\
Sketch        & 43.89           & 50.31   & 51.12 & 37.89  & 47.86          			&  50.38  & 54.07   &\textbf{58.58}        & 57.51      \\ \hline
Ave.\%        & 58.74           & 64.29   & 68.00 & 58.99   & 64.48         			&  64.65  & 67.05   &67.37			       & \textbf{69.21}      \\ \hline
\end{tabular}
}\vspace{-0.1cm}
\caption{Evaluation \% of classification on PACS. Multi-class accuracy (\%).}
\label{tab:acps4domain}
\end{table*}

\begin{table}[!h]
\centering
\scalebox{0.72}{
\begin{tabular}{ccccc}
\hline
\multirow{2}{*}{Unseen domain}  & \multicolumn{4}{c}{Ablation Study}  \\
		          & Tuning-Last& 2HE-Last & 2HE+Decom-Last & Ours-Full \\ \hline
Art painting      &  59.79      &    59.20      &  62.71  &   62.86           \\
Cartoon           &  56.22      &    55.50      &  52.69  &	  66.97  \\
Photo             &  86.79      &    87.33      &  88.84  &	  89.50  \\
Sketch            &  46.41      &	 48.45     &  	52.16  &  57.51  \\ \hline 
Ave.\%            &  62.30      &    62.62	  & 	64.10  &  69.21  \\ \hline
\end{tabular}
}\vspace{-0.1cm}
\caption{Ablation study. Multi-class accuracy (\%).}
\label{tab:acps3domain}
\end{table}


\subsubsection{Our PACS benchmark}
We compare baselines (SVM, 1HNN) and prior methods (LRE-SVM \cite{xu2014exploiting}, D-MTAE+1HNN \cite{ghifary2015domain}, uDICA \cite{muandet2013domainGen}) using DECAF$_7$ features against Deep-ALL, DSN \cite{bousmalis2016domain} and Ours-Full using end-to-end learning. From the results in Table~\ref{tab:acps4domain} we make the observations: (i) uDICA and D-MTAE-1HNN are the best prior DG models, and DSN is also effective despite being designed for DA. While uDICA scores well overall, this is mostly due to very high performance on the photo domain. This is understandable as in that condition DICA uses unaltered DECAF$_7$ features tuned for photo recognition. It is also the least useful direction for DG, as photos are already abundant.
(ii) As for the VLCS benchmark, Deep-ALL again performs well. (iii) However Ours-Full performs best overall by combining the robustness of a CNN architecture with an explicit DG mechanism.

\noindent\textbf{Ablation Study:}\quad To investigate the contributions of each components in our framework, we compare the following variants: \emph{Tuning-Last}: Trains on all sources followed by direct application to the target. But fine-tunes the final FC layer only. \emph{2HE-Last}: Fine-tunes the final FC layer, and uses our tensor weight generation (Eq.~\ref{eq:fc-decom}) based on 2-hot encoding for multidomain learning, before transferring the shared model component to the target. But without low rank factorisation. \emph{2HE+Decomp-Last}: Uses 2-hot encoding based weight synthesis, and low-rank decomposition of the final layer (Eq.~\ref{eq:fc-decom}). \emph{Ours-Full}: Uses 2-hot encoding and low-rank modeling on every layer in the CNN. 

From the results, we can see that each component helps: (i) 2HE-Last outperforms Tuning-Last, demonstrating the ability of our tensor weight generator to synthesize domain agnostic models for a multiclass classifier. (ii) 2HE+Decomp-Last outperforms 2HE-Last, demonstrating the value of our low-rank tensor modeling of the weight generator parameters. (iii) Ours-Full outperforms 2HE+Decomp-Last, demonstrating the value of performing these DG strategies at every layer of the network. 



\subsection{Further Analysis}
\noindent\textbf{Learned Layer-wise Sharing Strength}\quad 
An interesting property of our approach is that, unlike some other deep learning methods \cite{ganin2015udaBackprop,long2015deepAdaptationNetworks} it does not require manual specification of the cross-domain sharing structure at each layer of the CNN; and unlike Undo Bias \cite{ECCV12_Khosla} it can choose how to share more flexibly through the rank choice at each layer. We can observe the estimated sharing structure at each layer by performing Tucker decomposition to factorize the tuned model under a specified reconstruction error threshold ($\epsilon=0.001$). The resulting domain-rank at each layer reveals the sharing strength. 
The rank per-layer for each held-out domain in PACS is shown in Fig.~\ref{fig:sharing-strength}. Here there are three training domains, so the maximum rank is 3 and the minimum rank is 1. Intuitively, the results show heavily shared Conv1-Conv3 layers, and low-sharing in FC6-FC8 layers. The middle layers Conv4 and Conv5 have different sharing strength according to which domains provide the source set. For example, in Conv 5, when Sketch is unseen, the other domains are relatively similar so can have greater sharing, compared to when Sketch is included as a seen domain. This is intuitive as Sketch is the most different from the other three domains. This flexible ability to determine sharing strength is a key property of our model. 

\vspace{0.1cm}\noindent\textbf{Visualization}\quad To visualize the preferences of our multi-domain network, we apply the DGN-AM \cite{DBLP:journals/corr/NguyenDYBC16} method to synthesize the preferred input images for our model when parameterized (via the domain descriptor $z$) to one specific domain versus the abstract domain-agnostic factor. 
This visualization is imperfect because  \cite{DBLP:journals/corr/NguyenDYBC16} is trained using a photo-domain, and most of our domains are non-photographic art. 
Nevertheless, from  Fig.~\ref{fig:visualization4domain} the synthesis for Photo  domain seem to be the most concrete, while the Sketch/Cartoon/Painting domains are more abstract. 

\section{Conclusion} We presented a new dataset and deep learning-based method for domain generalization. Our PACS (Photo-Art-Cartoon-Sketch) dataset is aligned with a practical application of domain generalization, and we showed it has more challenging domain shift than prior datasets, making it suitable to drive the field in future. Our new domain generalization method integrates the idea of learning a domain-agnostic classifier with a robust deep learning approach for end-to-end learning of domain generalization. The result performs comparably or better than prior approaches.



{\small

}

\end{document}